%% file: ICDM_paper.tex
\documentclass{article}

\usepackage[final,nonatbib]{neurips_2018}

\usepackage[utf8]{inputenc} 
\usepackage[T1]{fontenc}    
\usepackage{hyperref}       
\usepackage{url}            
\usepackage{booktabs}       
\usepackage{amsfonts}       
\usepackage{nicefrac}       
\usepackage{microtype}      
\usepackage{amsmath,amssymb,amsfonts}
\usepackage{algorithmic}
\usepackage{algorithm}
\usepackage{subcaption}
\usepackage{graphicx}

\input{command_templates}

\newcommand{\ncl}{c}
\newcommand{\ny}{{\vec{x}}}
\newcommand{\nY}{\mathbf{X}}
\newcommand{\nxs}{\gamma}
\newcommand{\nx}{{\vec{\gamma}}}
\newcommand{\nz}{{\vec{z}}}
\newcommand{\nzs}{z}
\newcommand{\nX}{\mathbf{\Gamma}}
\newcommand{\nD}{\mathbf{D}}

\newcommand{\nA}{\mathbf{U}}
\newcommand{\na}{{\vec{u}}}
\newcommand{\nas}{u}
\newcommand{\nT}{T_0}
\newcommand{\nhs}{l}
\newcommand{\nh}{{\vec{l}}}
\newcommand{\nH}{\mathbf{L}}
\newcommand{\phiy}{\Phi(\nY)}
\newcommand{\phiys}{\Phi(\ny)}

\title{Confident Kernel Sparse Coding and\\ Dictionary Learning}

\author{%
  Babak Hosseini
  \thanks{
  	Preprint of the publication~\cite{hosseini2018conf}, as provided by the authors.
  	The final publication is available at IEEE Xplore via \url{https://ieeexplore.ieee.org/document/8594939}  	
  	} \\
	CITEC cluster of excellence\\
	Bielefeld University, Germany\\
  \texttt{bhosseini@techfak.uni-bielefeld.de} \\
   \And
   Barbara Hammer\\
	CITEC cluster of excellence\\
	Bielefeld University, Germany\\
   \texttt{bhammer@techfak.uni-bielefeld.de} \\
}

\begin{document}

\maketitle

\begin{abstract}
  In recent years, kernel-based sparse coding (K-SRC) has received particular attention due to its efficient representation of nonlinear data structures in the feature space.
  Nevertheless, the existing K-SRC methods suffer from the lack of consistency between their training and test optimization frameworks.
  In this work, we propose a novel confident K-SRC and dictionary learning algorithm (CKSC) which 
  focuses on the discriminative reconstruction of the data based on its representation in the kernel space. 
  CKSC focuses on reconstructing each data sample via weighted contributions which are confident in its corresponding class of data.
  We employ novel discriminative terms to apply this scheme to both training and test frameworks in our algorithm. 
  This specif design increases the consistency of these optimization frameworks and improves the discriminative performance in the recall phase.
  In addition, CKSC directly employs the supervised information in its dictionary learning framework to enhance the discriminative structure of the dictionary.
  For empirical evaluations, we implement our CKSC algorithm on multivariate time-series benchmarks such as DynTex++ and UTKinect.
  Our claims regarding the superior performance of the proposed algorithm are justified throughout comparing its classification results to the state-of-the-art K-SRC algorithms.
\end{abstract}


\input{intro_motive}
\input{discriminant_SC}
\input{proposed_alg}
\input{optimization_icdm}

\input{experiments}
\input{conclusion}

\subsubsection*{Acknowledgments}

This research was supported by the Cluster of Excellence Cognitive 
Interaction Technology 'CITEC' (EXC 277) at Bielefeld University, which
is funded by the German Research Foundation (DFG).

\bibliographystyle{IEEEtran}
\bibliography{IEEEabrv,/vol/semanticma/Thesis/Publications/Ref4Papers_CS}

\end{document}

%% file: command_templates.tex
\newcommand{\K}{{\mathcal K}}
\newcommand{\MB}[1]{\mathbf{#1}}
\newcommand{\MBB}[1]{\mathbb{#1}}
\newcommand{\MC}[1]{\mathcal{#1}}

\newcommand{\Tr}[0]{\mathrm{Tr}}

\newcommand{\T}{\top}

\newcommand{\smalleqb}[1]
{
	\begingroup
	\makeatletter
	\def
	\f@size{1}
	{#1}
    \endgroup
}

%% file: intro_motive.tex
\section{Introduction}
SRC algorithms try to construct the input signals using weighted combinations of few selected entries from a set of learned prototypes. 
The vector of weighting coefficients and the set of prototypes are called the \textit{sparse codes} and the \textit{dictionary} respectively \cite{Aharon2006}. 
Thus, the resulting sparse representation can capture the essential intrinsic characteristics of the dataset \cite{kim2010sparse}.
Based on that property, discriminant SRC algorithms are proposed
\cite{mairal2012task,liu2015joint} to learn a dictionary which can provide a discriminative representation of data classes. 
\cite{zhang2011image} showed that non-negative formulation of SRC could increase the possibility of relating each input signal to other resources from its class leading to a better classification accuracy.

Based on an implicit mapping of the data to a high-dimensional feature space, it is possible to 
formulate the kernel-based sparse coding (K-SRC)
which notably enhances the reconstruction and discriminative abilities of the SRC framework \cite{VanNguyen2013,bahrampour2015kernel}.
Nevertheless, the existing discriminant K-SRC models
suffer from the lack of consistency between their training and the recall optimization models. 
The supervised information plays a crucial role in efficiently approximating the sparse codes in their training phase. Nevertheless, the lack of such information in the recall phase 
degrades the discriminative quality of the sparse coding model for test data. 
\subsection{Our Contributions}
In this paper, we design a novel kernel-based sparse coding and dictionary learning for discriminative representation of the data. Our confident kernel sparse coding (CKSC) algorithm focuses on the
discriminative reconstruction of each data point using other data inputs which are taken mostly from one class.
Its kernel-SRC model is designed based upon a non-negative framework which facilitates such an intended representation. 
We emphasize the following contributions of our work which make it distinct from the relevant part of literature:
\begin{itemize}		
\item We employ the supervised information in training and test part of our CKSC algorithm, and its recall phase has a consistent structure with its training model, which enhances its performance regarding the test data.

\item The proposed non-negative discriminative sparse coding can learn each dictionary atom via constructing it upon mostly one class of data, but it can still be flexible enough to make small connections to other classes. 

\item Our discriminative dictionary learning method directly involves the supervised information in the optimization of both the dictionary and the sparse codes which make it more efficient regarding their discriminative objectives.


\end{itemize}

We review the relevant discriminative SRC methods in Section 2,
and our proposed CKSC algorithm is introduced in Section 3.
Its optimization steps are explained in Section 4,
and the experiments and results are presented in Section 5. Lastly,
the conclusion of the work is made in Section 6.

%% file: discriminant_SC.tex
\section{Background and Related Work}\label{sec_rel}
In this section, we briefly study relevant topics such as discriminant SRC and Kernel-based SRC.
\subsection{Discriminative Sparse Coding}
The training data matrix is denoted by $\nY=\left[\ny_1,...,\ny_N\right] \in \mathbb{R}^{d\times N}$, and we assume that the label information corresponding to $\nY$ is given as
$\nH=[\nh_1,\dots,\nh_N] \in\mathbb{R}^{p \times N}$, 
where $\nh_i$ is a binary vector such that $\nhs_{iq}=1$ if $\ny_i \in \{class~q\}$.
Discriminative sparse coding
is the idea of approximating each input signal as $\ny_i\approx \nD\nx_i$ while the cardinality norm $\|\nX\|_0$ is as small as possible. 
The matrices $\nD \in \mathbb{R}^{d\times c}$ and $\nX=[\nx_1,\dots,\nx_N] \in \mathbb{R}^{c\times N}$ are the dictionary and sparse codes respectively.
It is expected that the obtained $\nX$ also respects the supervised information $\nH$, such that each $\nh_i$ can be predicted based on the corresponding $\nx_i$.

Typical discriminative sparse coding algorithms such as \cite{mairal2012task,liu2015joint,quan2016supervised} focus on optimization problems generally similar to that of~\cite{mairal2012task} as
\begin{equation}
\begin{array}{ll}
\underset{\nX,\nD,\MB{W}}{\min}& \| \nY-\nD\nX\|_F^2+\alpha f(\nH,\MB{W},\nA,\nX)+\lambda \|\nX\|_1\\
\end{array}
\label{eq:disc_obj}
\end{equation}
where $\|.\|_F$ refers to the Frobenius norm, and $l_1$-norm $\|\nX\|_1$ is the relaxation of the $l_0$-norm which is advised by \cite{tibshirani1996regression}. 
The objective $\| \nY-\nD\nX\|_F^2$ measures the quality of the reconstruction of $\nY$ by dictionary $\MB{D}$.
Objective function $f(.)$ reflects the classification error, and
$\MB{W}$ denotes its matrix of parameters.
The scalars $\alpha, \lambda$ specify the weights of the sparsity and discriminative terms respectively. 
In the framework of (\ref{eq:disc_obj}), the objective function $f(.)$ is considered to be jointly in terms of ($\MB{D},\nX,\nH$). Nevertheless, in typical examples of discriminative SRC methods
\cite{mairal2012task,liu2015joint}, the update of $\nX$ or $\nD$ happens disjointed from the supervised information $\nH$. Consequently, this often leads to low-quality local convergence points. 

In some of the sparse coding works similar to 
\cite{bach2008learning,jenatton2010proximal,yang2011fisher,kong2012dictionary}, 
the dictionary $\nD$ is split into multiple class-specific sub-dictionaries which are separately trained to reconstruct each class of data. 
In \cite{kong2012dictionary} 
they learn a common dictionary module which is shared among all of the classes.
Nevertheless, these methods usually face problems when different data classes are close to each other, 
and dictionary atoms from another class can also express some data points from one class.
%
%


Generally, a sparse coding framework which can incorporate the supervised training information in the recall phase provides a more efficient discriminant mapping for the test data.
To our knowledge, the only discriminative sparse coding algorithm which merely aims for such consistency is the Fisher Discriminant sparse coding \cite{yang2011fisher} which tries to reconstruct encodings of test data close to the average value of all $\nx_i$ related to the presumed class. However, in contrast to its base assumption, it is convenient for an SRC model to obtain distributed clusters of sparse codes, even though they are related to one class. 
\subsection{Kernel-based SRC}
Incorporating kernel representation into sparse coding can extend it to 
nonlinear and non-vectorial domains \cite{VanNguyen2013,hosseini2016non}.
Accordingly, $\Phi: \mathbb{R}^{d} \rightarrow \mathbb{R}^h$ denotes as an implicit non-linear mapping which can transfer data to a reproducing kernel Hilbert space (RKHS). 
Therefore, we can use the kernel function $\K(\nY,\nY)$ in the input space which is associated with the implicit mapping $\Phi$ such that $\K(\nY,\nY)=\langle \Phi(\nY),\Phi(\nY)\rangle$. 
Throughout using the kernel representation in the feature space, SRC can be reformulated similar to: 
\begin{equation}
\underset{\nX,\nD}{\min}\| \Phi(\nY)-\Phi(\nD)\nX\|_F^2 \quad \mathrm{s.t.}~\|\nx_i\|_0 < \nT ~~\forall i
\label{eq:sprs_ker}
\end{equation}
in which $\Phi(\nD)$ is the dictionary defined in the feature space. 


It is shown by \cite{VanNguyen2013} that it is possible to define a dictionary in the feature space in the form of $\Phi(\nD)=\Phi(\nY)\nA$ where $\nA \in \mathbb{R}^{N\times c}$. 
Each column of the dictionary matrix $\nA$ contains a linear combination of data points from the feature space.
However, to its advantage, 
the reconstruction term in (\ref{eq:sprs_ker}) can be rephrased in terms of the Gram matrix $\K(\nY,\nY)$ of the given data 
\begin{equation}
\begin{array}{l}
\| \Phi(\nY)-\Phi(\nY)\nA\nX\|_F^2=\\
\K(\nY,\nY)+\nX^\top \nA^\top \K(\nY,\nY) \nA \nX- 2\K(\nY,\nY)\nA\nX
\end{array}
\label{eq:orec}
\end{equation}
which facilitates optimizing of dictionary matrix $\nA$.


%% file: proposed_alg.tex
\section{Confident Kernel Sparse Coding and Dictionary Learning}\label{sec:disc}
We propose a novel kernel-based discriminative sparse coding algorithm with the following training framework 


\begin{equation}
\begin{array}{lll}
{\footnotesize Train:}&
\underset{\nX,\nA}{\min}
& \|\Phi(\nY)-\Phi(\nY) \nA \nX\|_F^2
+\alpha \MC{F}(\nH,\nX,\nA) \\
&\mathrm{s.t.} & \|\nx_i\|_0 < \nT,
~~\|\phiy\na_i\|_2^2=1,\\
&& \|\na_i\|_0 \le \nT,~~ \nas_{ij}, \nxs_{ij} \in \mathbb{R}^{+},~~\forall ij\\
\end{array}
\label{eq:train}
\end{equation}
and its relevant recall framework as
\begin{equation}
\begin{array}{lll}
Recall:&
\underset{\nx}{\min}
& \|\Phi(\nz)-\Phi(\nY) \nA \nx\|_F^2
+\alpha \MC{G}(\nH,\nx,\nA) \\
&\mathrm{s.t.} & \|\nx\|_0 < \nT,
~\nxs_i\in \mathbb{R}^{+}~~\forall i\\
\end{array}
\label{eq:test}
\end{equation}
Similar to the work of \cite{VanNguyen2013}, $\nX$ and $\nA$ denote sparse codes and the dictionary matrix of this framework respectively.
The first part of the objectives in (\ref{eq:all_train}) and (\ref{eq:all_test})) 
is the reconstruction loss function used in (\ref{eq:orec}),
and $\{\MC{F},\MC{G}\}$ are the novel discriminative loss terms we introduce in this paper. 
Parameter $\nT$ applies the $l_0$-norm sparsity constraint on the columns of $\{\nA,\nX\}$, and $\alpha$ is the control factor between the reconstruction and the discriminant terms. 

Furthermore, we can write $\phiy \nA \nx =\phiy \vec{s}$ where $\vec{s} \in \MBB{R}^N$.
Therefore, by using the dictionary structure of $\Phi(\nY)\nA$, it is also necessary to bound the value of $\|\na_i\|_0$.
That way, each $\phiys$ is encoded by selecting a few contributions from the training set $\nY$ (small $\|\vec{s}\|_0$).
Furthermore, the constraint $\|\phiy \na_i\|_2^2=1$ is a bound on $l_2$-norm of the dictionary columns as a convenient way to prevent the solution of \eqref{eq:train} from becoming degenerated \cite{Aharon2006}.

In the following sub-sections, we discuss the mathematical detail of the novel objective terms employed in (\ref{eq:all_train}) and (\ref{eq:all_test}) and explain the motivations behind our such choice of design.

\subsection{Discriminative Objective $\MC{F}(\nH,\nA,\nX)$:} \label{sec:dis_rep}
Before discussing the mathematical content of $\MC{F}(\nH,\nA,\nX)$, we like to explain the motivation behind our specific choice of $\MC{F}$ as the discriminant term. 
If $\Phi(\ny)$ is reconstructed as $\phiys=\phiy\nA\nx$, then the entries of $\nH\nA\nx \in \MC{R}^\ncl$ show the share of each class in the reconstruction of $\phiys$. 
Accordingly, we denote each $s$-th row of the labeling matrix $\nH$ as $\vec{\rho}^\top_s$ such that $\rho_{si}=1$ if $\ny_i$ belongs to class $s$.
Now, as an extreme case, if we assume that $\ny$ belongs to the class $q$ and $\Phi(\ny)$ is lying on the subspace of class $q$ in the feature space, we have $\vec{\rho}_s^\top\nA\nx=0~\forall s\neq q$.
By generalizes this extreme case to a more realistic condition we define  
$\MC{F}(\nH,\nA,\nx)=\underset{s \neq q}{\sum} \rho_s^\top \nA \nx$
as the sum of contributions from other classes (than $q$). 
Hence, for all $\nX$ we have
\begin{equation}
\MC{F}(\nH,\nA,\nX)=\Tr((\MB{1}-\nH^\top) \nH \nA \nX )
\label{eq:disct_tr}
\end{equation}
where $\MB{1} \in \MBB{R}^{N \times \ncl}$ is a matrix of all-ones, and $\Tr(.)$ denotes the matrix trace.
Function $\MC{F}(\nH,\nA,\nX)$ is a linear term with respect to each $\nx_i$ and $\na_i$ individually. 
Therefore, it does not violate the convexity of the optimization problem in (\ref{eq:train}).
Considering the optimization framework of (\ref{eq:all_train}), $\MC{F}$ is employed along with the additional term $\beta \|\nA \nX\|_F^2$.
This term preserves the consistency between the training and the recall models and is explained in the next subsection.

\subsection{Discriminative Recall Term $\MC{G}(\nH,\nx,\nA)$:} \label{sec:ent_recall}
For a test vector $\nz$, 
we assume that $\Phi(\nz) \in span \{\phiy\}$ and belongs to the class $q$ such that its projection on subspace $q$ as $\|\Phi(\nz)^q\|_2$ is arbitrarily larger than $\|\Phi(\nz)\|_2-\|\Phi(\nz)^q\|_2$.
Therefore, 
via using the learned $\nA$ from Eq \eqref{eq:train}, there exists a $\nx$ which reconstructs the test data as $\Phi(\vec{z})=\phiy \nA \nx$ with more contributions chosen from the class $q$.
Consequently, 
the class label $\nh_{\nzs}$ is predicted as $\nhs_{\nzs}(j)=1$ where
\begin{equation}
j=\underset{j}{\text{argmax }} \vec{\rho}^\top_j \nA \nx \\
\end{equation}
In other words, $\nh_{\nzs}$ is determined as the class of data which has the most contribution to the reconstruction of $\nz$.

Since we do not have access to the labeling information for the test data, we propose a cross-entropy-like loss for (\ref{eq:test}) as
\begin{equation}
\MC{G}(\nH,\nx,\nA)=\sum_i(\sum_{s\neq i}h_s)h_i \quad \text{  where } h_i:=\rho_i^\top \nA \nx
\label{eq:g_recal} 
\end{equation}
Since $\{\nx,\nA\}$ are non-negative matrices, $\MC{G}$ is non-negative as well and can have its global optima where $\MC{G}(\nx^*)=0$. 
Denoting $\vec{h}^* = \nH\nA\nx^*$, besides the trivial point of $\vec{h}^*=0$, 
the loss term reaches its global optima when $\vec{h}^*$ contains only one non-zero value in its $i$-th entry.
This observation is equivalent to finding $\nx^*$ such that it
reconstructs $\nz$ using contributions only from one class of data.
Consequently, the non-trivial minima of both regularization terms in (\ref{eq:disct_tr}) and (\ref{eq:g_recal}) occur at similar points where the decision vector $\nH \nA \nx$ has approximately a crisp form regarding only one of its entries. Therefore, adding $\MC{G}$ increases the consistency between training and the test frameworks.\\

By re-writing $\MC{G}(\nH,\nx,\nA)=\nx^\T \nA^\T  \nH^\T (\MB{1}-\MB{I}) \nH \nA \nx$,
it is possible to show that $\MC{G}$ is a non-convex function due to the existing term 
$(\MB{1}-\MB{I})$ in its formulation.
To fix this issue, we add $\beta\|\nA\nx\|_2^2$ to the objective of (\ref{eq:test}) which converts its 
second-degree terms to  
$$\nx^\top \nA ^\top (\MB{V}+\beta \MB{I}_{N \times N}) \nA \nx$$ 
where $\MB{V}:=\K(\nY,\nY) + \alpha\nH^\top(\MB{1}-\MB{I}_{\ncl \times \ncl}) \nH$.		
Now, denoting $\{\lambda_i\}_{i=1}^N$ as the eigenvalues of $\MB{V}$, we choose 
$\beta=-\min_i \lambda_i$ makes $(\MB{V}+\beta \MB{I}_{N \times N})$ a positive semi-definite matrix (PSD) and consequently, the whole objective becomes PSD due to its quadratic form. Therefore, (\ref{eq:test}) becomes a convex problem by adding this term.   

In order to preserve the consistency between the test and training model, we also add the term $\beta\|\nA \nX\|_F^2$ to the discriminant loss $\MC{F}$ of (\ref{eq:disct_tr}) which results in (\ref{eq:all_train}).
Doing so, we want to make sure the trained dictionary $\nA$ has a proper structure also regarding (\ref{eq:test}).	
Furthermore, parameter $\beta$ is independent of the test data and is computed only once before the start of the optimization phase. 
In the next section, we explain the optimization steps regarding frameworks of (\ref{eq:train}) and (\ref{eq:test}). 
%
%
%

%
%

%% file: optimization_icdm.tex
\section{Optimization Scheme}
By re-writing (\ref{eq:train}) and (\ref{eq:test}) using the provided descriptions of $\MC{F}$ and $\MC{G}$ in section \ref{sec:disc}, we obtain the following training optimization framework
\begin{equation}
\begin{array}{lll}
Train: ~~&
\underset{\nX,\nA}{\min}&  \|\phiy-\phiy \nA \nX\|_F^2
+\beta \|\nA \nX\|_F^2\\ 
&&+\alpha \Tr\{ (\MB{1}-\nH^\top) \nH \nA \nX \}\\
&\mathrm{s.t.} & \|\nx_i\|_0 < \nT,
~~\|\phiy\na_i\|_2^2=1,\\
&& \|\na_i\|_0 \le \nT,~~ \nas_{ij}, \nxs_{ij} \in \mathbb{R}^{+},~~\forall ij\\
\end{array}
\label{eq:all_train}
\end{equation}
and its relevant recall problem as
\begin{equation}
\begin{array}{lll}
Test: ~~&
\underset{\nx}{\min}& \|\Phi(\nz)-\phiy \nA \nX\|_F^2
+\beta \|\nA \nX\|_F^2\\
&&+\alpha (\nx^\T \nA^\T  \nH^\T (\MB{1}-\MB{I}) \nH \nA \nx) \\
&\mathrm{s.t.} & \|\nx\|_0 < \nT,
~\nxs_i\in \mathbb{R}^{+}~~\forall i\\
\end{array}
\label{eq:all_test}
\end{equation}
Although the optimization problem (\ref{eq:all_train}) is not convex w.r.t $\{\nA,\nX\}$ together, We train the discriminant sparse coding model in 2 alternating convex optimization steps. At each step, we update one of the parameters while fixing the other one as presented in Algorithm \ref{alg:whole}. 


\subsection{Update of the Sparse Codes $\nX$} 
The entire objective function in (\ref{eq:all_train}) has a separable column structure with respect to $\nX$, and it can be optimized for each $\nx_i$ individually.
Therefore, after removing the constant terms, (\ref{eq:all_train}) is re-written w.r.t each $\nx_i$ as 
\begin{equation}
\begin{array}{ll}
\underset{\nx_i}{\min}&  \nx_i^\top\big[\nA^\top(\K + \beta \MB{I}) \nA \big] \nx_i\\
&+\big[ \alpha (\MB{1}-\nh_i^\top) \nH \nA -2 \K(\ny_i,\nY) \nA  \big] \nx_i \\
\mathrm{s.t.} & \|\nx_i\|_0 < \nT,~~ \nxs_{ij} \in \mathbb{R}^{+},~~\forall ij\\
\end{array}
\label{eq:qp_x_1}
\end{equation}
where $\K$ stands for $\K(\nY,\nY)$.
This optimization framework is a non-negative quadratic programming problem with the constraint $\|\nx_i\|<\nT~\forall i$. Furthermore, $\K+\beta \MB{I}$ is a PSD matrix, and consequently (\ref{eq:qp_x_1}) is a convex problem.

Therefore, we can employ 
the Non-Negative Quadratic Pursuit (NQP) algorithm \cite{hosseini2018confidentjmlr} to optimize (\ref{eq:qp_x_1}). 
NQP algorithm generalizes the Matching Pursuit approach \cite{Aharon2006} for quadratic problems similar to (\ref{eq:qp_x_1}).
\subsection{Update of the Dictionary $\nA$} 
Similar to (\ref{eq:qp_x_1}), it is also possible to reformulate the objective terms of (\ref{eq:all_train}) w.r.t. each dictionary column $\na_i$ separately. Accordingly, the reconstruction loss in (\ref{eq:all_train}) is equivalent to 
\begin{equation}
\begin{array}{l}
\| {\Phi}(\nY) \mathbf{E}_i-{\Phi}(\nY)\na_i \nx^i\|_F^2, 
~~ \MB{E}_i=(\mathbf{I}-\underset{j\neq i}{\sum} \na_j \nx^j) 
\end{array}
\label{eq:ei}
\end{equation}
in which $\nx^i$ is the $i$-th row of $\nX$. Likewise, the rest of the objective term in (\ref{eq:all_train}) can be written in terms of $\na_i$ as 
$$\beta {\na_i}^\top(\nx^i {\nx^{i\top}} \MB{I})\na_i
+\alpha \nx^i (\MB{1}-\nH^\T)\MB{L}\na_i 
+2\beta \nx^i (\MB{I}-\MB{E}_i^\T) \na_i$$
where the constant parts are eliminated.
So, using the kernel function $\K(\nY,\nY)$, we re-formulate (\ref{eq:all_train}) for updating $\na_i$ as
\begin{equation}
\begin{array}{ll}
\underset{\na_i}{\min}
&\beta {\na_i}^\top(\nx^i {\nx^{i\top}} (\K+\beta \MB{I}))\na_i\\
&+\nx^i \big[ \alpha (\MB{1}-\nH^\T)\MB{L} 
+2\beta (\MB{I}-\MB{E}_i^\T) 
-2 \MB{E}_i^\top \K \big] \na_i\\
\mathrm{s.t.} & \|\phiy \na_i\|_2^2=1,~ \|\na_i\|_0 \leq \nT ~,~  \nas_{ij}\in \mathbb{R}^{+}~~\forall j
\end{array}
\label{eq:up_u}
\end{equation}
Similar to (\ref{eq:qp_x_1}), the above framework has the non-negative quadratic form with the cardinality constraint $\|\na_i\|_0 \leq \nT$ and is a convex problem  similar to (\ref{eq:qp_x_1}). Consequently, its solution can be approximated using the NQP algorithm~\cite{hosseini2018confidentjmlr}.

\textbf{Note:} When computing $\MB{E}_i$ to update $\na_i$, matrix ${\nA}$ should be used with its recently updated columns to improve the convergence speed of the optimization loop. 
For example, in a successive update of columns in $\nA$ starting from the 1st column, we compute $\MB{E}_i$ at iteration $t$ by using matrix $\nA$ as:
$${\nA}=\{\na_1^{1},\na_2^{2},...,\na_{i-1}^{(t-1)},\na_{i}^1,...,\na_{k}^1\}$$
where $\na_i^t$ is the value of $\na_i$ at iteration $t$ of the dictionary updating loop.
Besides, directly after updating each $\na_i$ via NQP algorithm, it should be normalized as 
$ \na_i \rightarrow \frac{\na_i}{\|\Phi(\nY)\na_i\|_2}$.  



\begin{algorithm} 
	\caption{Confident Kernel Sparse Coding } 
	\label{alg:whole} 
\begin{algorithmic} 
\STATE {\bfseries Parameters:} $\lambda,\nT$.
\STATE {\bfseries Input:} Labels $\nH$, kernel matrix $\K(\nY,\nY)$.
\STATE {\bfseries Output:} Sparse coefficients $\nX$, discriminant dictionary $\nA$.
\STATE {\bfseries Initialization:} Computing $\beta$ based on section \ref{sec:ent_recall}.
\REPEAT 
\STATE 		Updating $\nX$ based on (\ref{eq:qp_x_1}) using NQP
\STATE 		Updating $\nA$ based on (\ref{eq:up_u}) using NQP
\UNTIL{Convergence}
\end{algorithmic}
\end{algorithm}	

\subsection{Update of the Recall Phase $\nx$:} 
In order to reconstruct the test data $\nz$, its corresponding sparse code $\nx$ is approximated 
via expanding (\ref{eq:all_test}) as follows
\begin{equation}
\begin{array}{ll}
\underset{\nx}{\min}& 
\nx^\top \nA ^\top \big[\K + \alpha\nH^\top(\MB{1}-\MB{I}) \nH+\beta \MB{I} \big] \nA \nx\\ 
&-2 \K(\nz,\nY) \nA \nx\\
\mathrm{s.t.} & \|\nx\|_0 < \nT,~~ \nxs_{j} \in \mathbb{R}^{+}~~\forall j\\
\end{array}
\label{eq:upz}
\end{equation}
This optimization problem is convex based on the analysis provided in Sec.~\ref{sec:ent_recall},
and can be approximately solved by the NQP algorithm \cite{hosseini2018confidentjmlr} similar to the update of $\nX$ and $\nA$.
The algorithm's code will be available in an online repository\footnote{https://github.com/bab-git/CKSC}.

%% file: experiments.tex
\section{Experiments}
In order to evaluate our proposed confident kernel sparse coding algorithm, we carry out implementations on the following selection of sequential datasets. 
\begin{itemize}
	\item Cricket Umpire\cite{ko2005online}: A collection of 180 instances of hand movement data related to 12 different types of cricket umpire's signal \cite{ko2005online}.
	\item Articulatory  Words\cite{wang2014preliminary}: Recorded data of facial movements using EMA sensors while uttering 25 different words, containing 575 data samples \cite{wang2014preliminary}.
	\item Schunk Dexterous \cite{drimus2014design}: Tactile data of robot grasping for 5 different objects with 10 samples per object. The dimensions are extracted based on the work of \cite{madry2014st}.
	\item UTKinect Actions
	 \cite{xia2012view}: 
	 3D location of body joints recorded using Kinect device related to 9 different actions, containing 199 action instances in total.
	\item DynTex++: A large-scale Dynamic Texture dataset with 36 classes and 100 sequences per category \cite{ghanem2010maximum}.
\end{itemize} 

For DynTex++, the kernel is computed based on \cite{quan2016equiangular}; however, for other datasets, we compute the Gaussian kernel 
$$\K(\ny_i,\ny_j)=exp(-\mathcal{D}(\ny_i,\ny_j)^2/\delta)$$ 
where $\mathcal{D}(\ny_i,\ny_j)$ is the distance between a pair of $\{\ny_i,\ny_j\}$. The 	value of $\mathcal{D}(\ny_i,\ny_j)$ is computed using Dynamic Time Warping method \cite{Adistambha2008}, and
$\delta$ is determined as the average of $\mathcal{D}(\ny_i,\ny_j)$ over all data points.

We compare our proposed method to the relevant baseline algorithms selected as
K-KSVD \cite{VanNguyen2013}, JKSRC \cite{liu2016object}, LC-NNKSC \cite{hosseini2016non}, LP-KSVD \cite{liu2015joint}, KGPL \cite{harandi2013dictionary}, and EKDL \cite{quan2016equiangular},   
which are known as the most recent kernel-based discriminative sparse coding algorithms.
The basis of our comparisons is the average classification accuracy for 10 randomized test/train selections, which shows the quality of the resulted discriminative representations for each method.
$$\text{accuracy}=\frac{\{\#\text{ correctly assigned labels}\}}{N}\times 100 $$
\textbf{Note:} It is important to emphasize that the purpose of our discriminative sparse coding framework is to 
obtain a \textbf{sparse discriminant representation} of the data based on its pre-computed \textbf{kernel representation}.
Therefore, instead of comparing the results to all the available top classifiers such as Deep Neural Networks and others, we only select the recent kernel-based alternatives which fit the above description. 

\textbf{Parameters Tuning:}
In order to tune the parameters $\lambda$ and $\nT$ related to the optimization framework of CKSC in (\ref{eq:train} , \ref{eq:test}), we also perform 5-fold cross-validation. We carry out the same procedure for the baselines to find their optimal choice of parameters. 
The parameter $k$ (dictionary size) is determined by choosing the \textit{number of dictionary atoms per class}: 
$$\{\# \text{ atoms per class}\} =\{\#\text{ classes} \} \times \nT$$
However, in practice as a working parameter setting for CKSC, we can choose a value around $\lambda=0.1$. 
Parameter $\nT$ (and the dictionary size) depends on class distributions and complexity of the dataset. However, based on practical evidence having large values for $\nT$ does not improve the performance of CKSC and only increases the dictionary redundancy.
Table \ref{tab:params1} the chosen parameter values for CKSC algorithm regarding each dataset.

\begin{table}
	\centering
	\caption{Parameter settings of CKSC regarding each dataset}	
	\begin{tabular}{  |c|c|c|c|c| c| } %
		\hline	
		Parameter & Cricket & Words & Schunk & UTKinect & DynTex++\\ 		
		\hline
		 $\nT$    & 6&8&5&7&15\\
		\hline
		$\lambda$ & 0.2&0.2&0.15&0.1&0.15\\
		\hline
	\end{tabular} 			
	\label{tab:params1} 
\end{table}
\subsection{Classification Results}
The classification accuracies of the implementations are reported in Table \ref{tab:kernel}. 
According to the given results, our CKSC algorithm obtains the highest classification performance for all of the benchmarks, which shows that the designed frameworks of (\ref{eq:train}) and (\ref{eq:test}) provide better discriminative representations in comparison to other K-SRC algorithms. 

CKSC, EKDL, LP-KSVD, and LC-NNKSC algorithms can be considered as the runner-up methods which have competitive classification accuracies. 
Their good performance is due to the embedded labeling information in their discriminant terms which improves their discriminative representations in contrast to K-KSVD, JSRC, and KGDL. 
Nevertheless, there is a variation in the comparison results of these methods. We can conclude that although they use different strategies to obtain a discriminant model, their recall model does not necessarily comply with their training model.
In contrast, CKSC demonstrated that it has a more efficient embedding of the supervised information in both of the training and the recall framework.

LC-NNKSC uses a non-negative framework similar to the basis of CKSC's structure, and 
via comparing the results of LC-NNKSC to those of LP-KSVD and EKDL we observe that its non-negative framework obtains a competitive performance. 
Although LP-KSVD and EKDL employ extra objective terms in their models,
this non-negative structure can achieve a similar outcome without the need to use such extra terms.
Relevantly, CKSC benefits from this non-negative optimization framework as a basis for its confidence based model which leads to its superior performance compared to other baselines.
\begin{table*}[h]
	\centering
	\caption{Average classification accuracies ($\%$) $\pm$ standard deviations for the selected datasets.}
		\resizebox{1\textwidth}{!}{%
	\begin{tabular}{|l|c| c|c| c| c| c| c| } 
		\hline				
		Datasets & K-KSVD & JKSRC & LC-NNKSC & LP-KSVD & KGDL &  EKDL & \textbf{CKSC} \\
		\hline	
		\hline	
		Schunk  &83.42$\pm$0.35&87.49$\pm$0.57&89.96$\pm$0.64&89.62$\pm$0.51&88.17$\pm$0.43&88.39$\pm$0.24&\textbf{91.42$\pm$0.34}\\
		\hline	   	
		DynTex++ &89.22$\pm$0.47&89.95$\pm$0.35&93.22$\pm$0.37&93.12$\pm$0.47&92.83$\pm$.31&93.51$\pm$0.46&\textbf{94.36$\pm$0.32}\\
		\hline	   	
		Words  &93.35$\pm$0.84&92.14$\pm$0.78&97.33$\pm$0.75&97.64$\pm$0.67&95.82$\pm$0.66&96.53$\pm$0.68&\textbf{98.82$\pm$0.69}\\
		
		\hline
		
		Cricket &75.12$\pm$1.54&78.14$\pm$1.38&83.33$\pm$1.12&83.33$\pm$0.95&82.78$\pm$1.07&84.25$\pm$0.95&\textbf{86.45$\pm$0.85}\\
		\hline				
		
		UTKinect &80.26$\pm$0.35&81.47$\pm$0.42&84.12$\pm$0.32&84.93$\pm$0.25&83.51$\pm$0.36&84.18$\pm$0.24&\textbf{86.52$\pm$0.26}\\
		\hline			
	\end{tabular}
}
	\\The best result (\textbf{bold}) is according to a two-valued t-test at a $5\%$ significance level.
	\label{tab:kernel}
\end{table*}
\subsection{Sensitivity to the Parameter Settings}
\begin{figure}[tb]
	\begin{subfigure}{0.48\textwidth}
		\centering		
		\includegraphics[width=.6\linewidth]{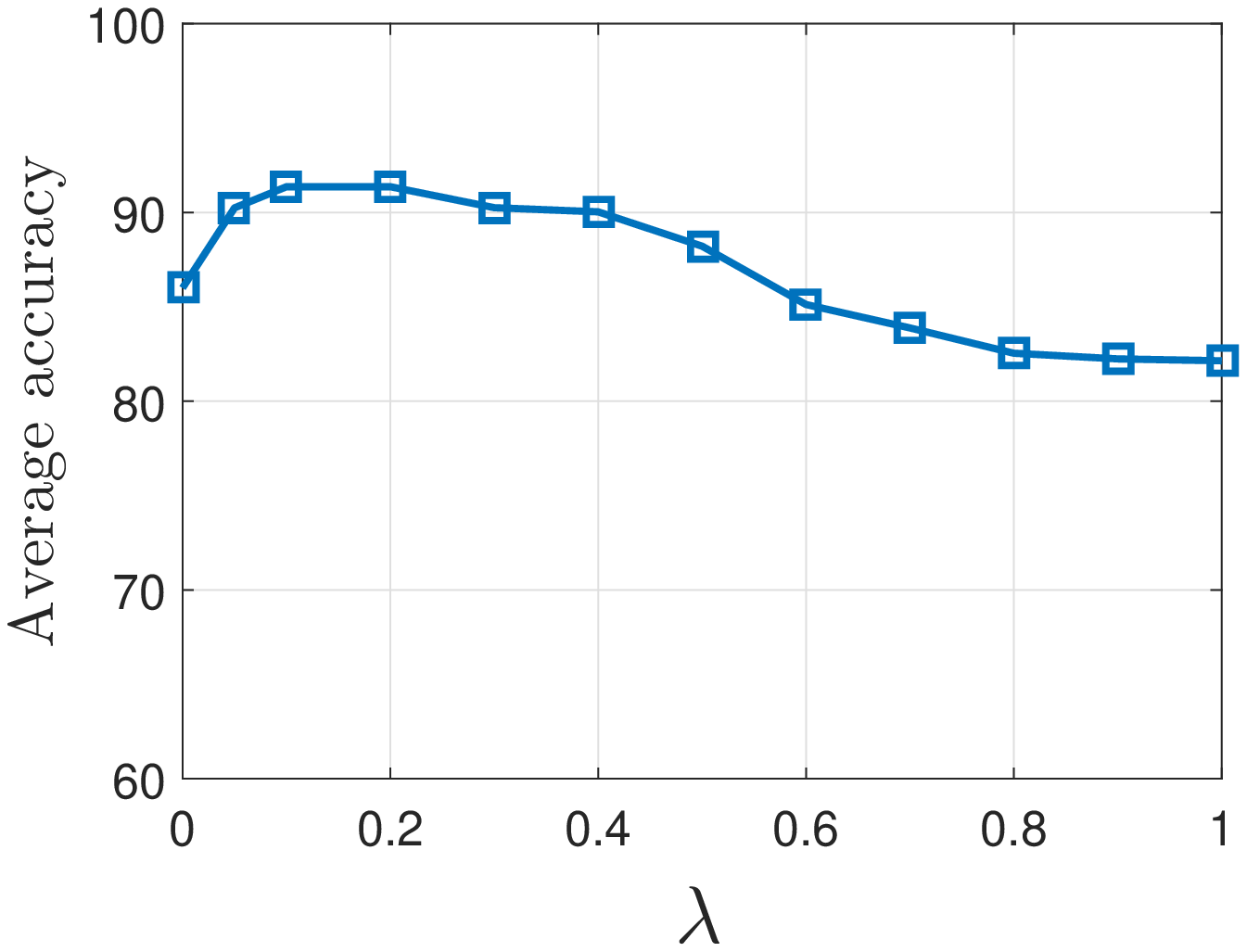}
		\caption{}
	\end{subfigure}	
	\begin{subfigure}{0.48\textwidth}
		\centering		
		\includegraphics[width=.6\linewidth]{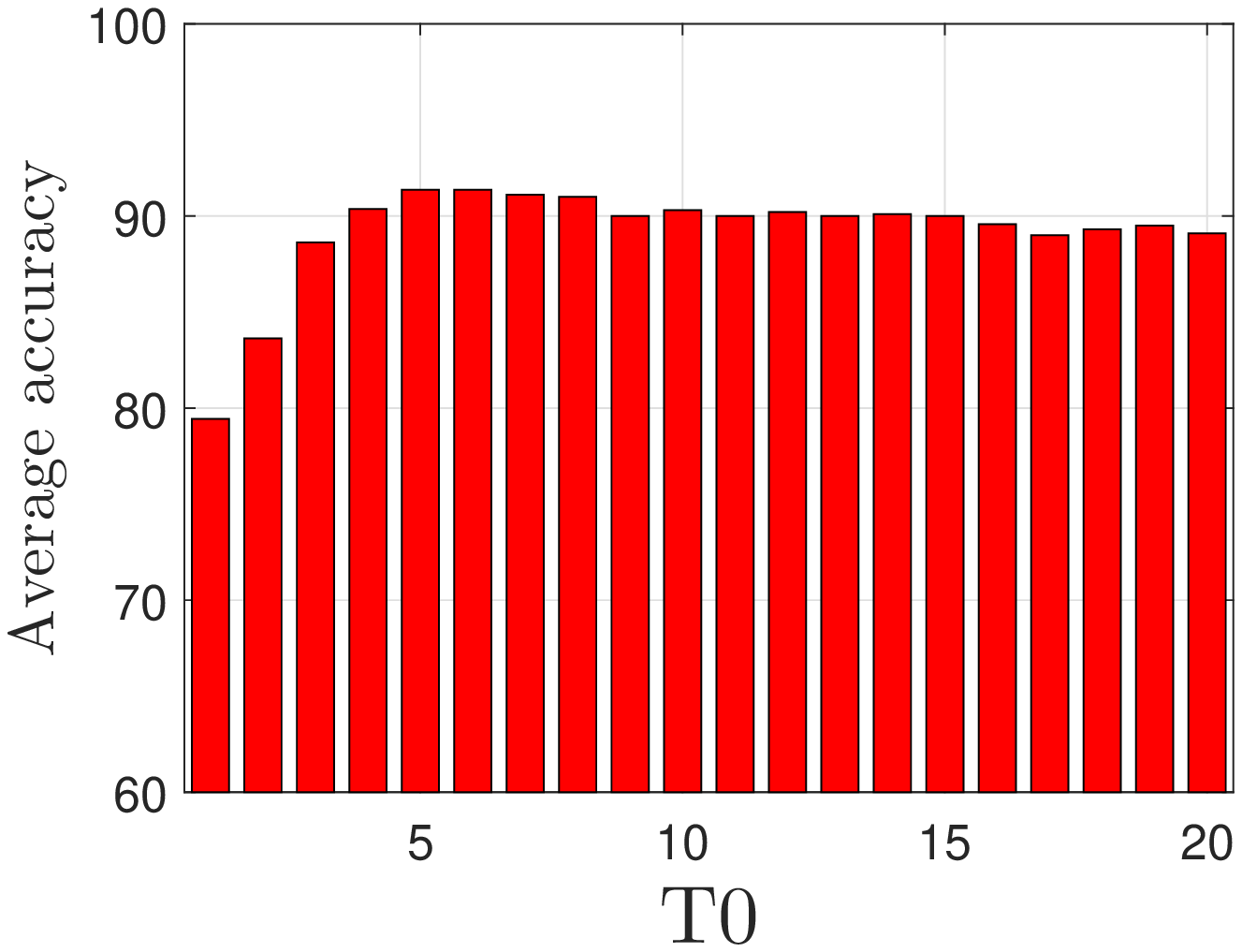}
		\caption{}			
	\end{subfigure}	
	\caption{Sensitivity analysis of CKSC to hyper-parameters (a)$\lambda$ and (b)$\nT$ for Schunk dataset.}
	\label{fig:sens}
\end{figure}

To study the sensitivity of CKSC to the parameter settings, we carry out experiments via changing the algorithm's parameters $(\lambda,\nT)$. Implementing on Schunk dataset, we apply CKSC in 2 individual settings via changing one parameter throughout each experiment when the other one is fixed to the value reported in Table \ref{tab:params1}. 
As it is observed from Figure \ref{fig:sens}a, a good choice for $\lambda$ lies in the interval $\left[0.1,0.4\right]$. However, the discriminative objective can outweigh the reconstruction part for values of $\lambda$ close to 1, and it results in over-fitting and performance reductions.

Regarding the dictionary size, we increase $\nT$ from 1 to 20 with step-size 1 which changes the size of $\nA$ in the range $\left[20,400\right]$ with step-size 20 (average number of data samples per class). According to Figure \ref{fig:sens}b, for Schunk dataset, having $\nT$ between 4 and 8 keeps the performance of CKSC at an optimal level. 
As it is clear, small values of $\nT$ put a tight limit on the number of available atoms $\na_i$ for reconstruction purpose which reduces the accuracy of the method.
On the other hand, larger values of $\nT$ increase dictionary redundancy and loosen up the sparseness bound on $\nX$; nevertheless, NQP algorithm and the non-negativity constraints intrinsically incur sparse characteristics to $\nX$ and $\nA$ via combining only the most similar resources. Therefore, increasing $\nT$ does not dramatically degrades the performance of CKSC. 

\subsection{Complexity and Convergence of CKSC}
In order to calculate the computational complexity of CKSC per iteration, we analyze the update of $\{\nX,\nA\}$ separately.
In each iteration, $\nX$ and $\nA$ are optimized using the NQP algorithm which has the computational complexity of $\MC{O}(N\nT^2)$ \cite{hosseini2018confidentjmlr}, 
where $\nT$ and $N$ are the sparsity limit and 
the number of training samples. 
Therefore, optimizing $\nX$ and $\nA$ cost 
$\MC{O}(kN\nT^2+kN^2)$ and $\MC{O}(kN\nT^2+kcN^2+kN^3)$ respectively, in which $c$ is the dictionary size. 

As shown in Figure \ref{fig:sens}b, practically we choose $\nT\leq 10$, and also $\ncl< k$. Therefore the total time complexity of each iteration is $\MC{O}(kN^3)$, in which, the dominant parts of the computational costs are dedicated to pre-optimization computations.
Nevertheless, for datasets that $N/\ncl$ is relevantly large, the size of $\nA$ should be chosen as $k \ll N$ in practice. Otherwise, it increases the redundancy in the dictionary without having any added-value. Therefore, in practice, the computational complexity of CKSC becomes $\MC{O}(N^3)$.

The optimization framework of CKSC in (\ref{eq:all_train}) is non-convex when considering $\{\nA,\nX\}$ together. However, each of the sub-problems defined in (\ref{eq:qp_x_1}) and (\ref{eq:up_u}) are convex. Therefore, the alternating optimization scheme in Algorithm \ref{alg:whole} converges in a limited number of steps. 

\begin{figure}
	\begin{subfigure}{0.48\textwidth}
		\centering		
		\includegraphics[width=0.83\linewidth]{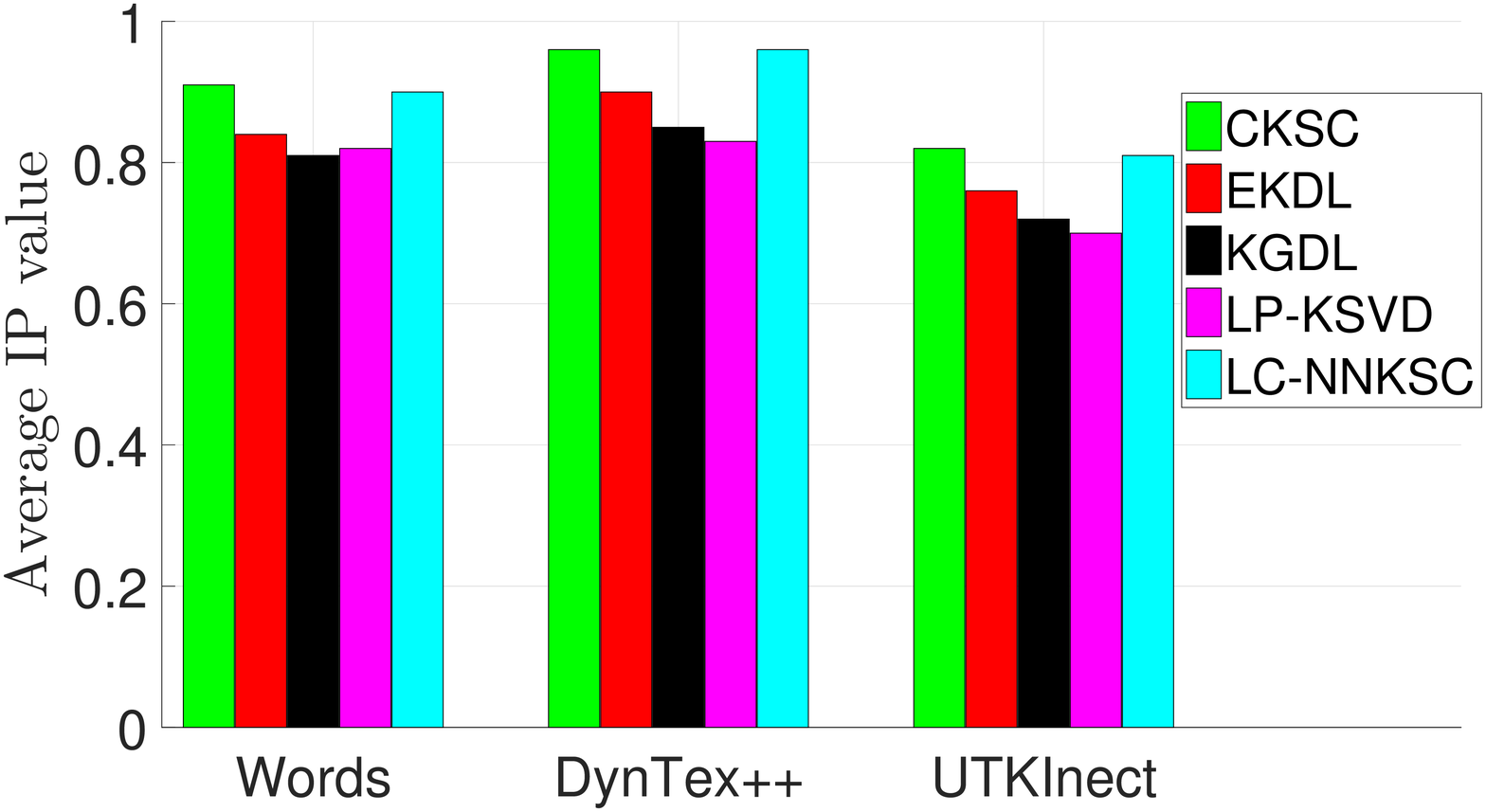}
		\caption{}		
	\end{subfigure}
	\begin{subfigure}{0.48\textwidth}
		\centering		
		\includegraphics[width=0.6\linewidth]{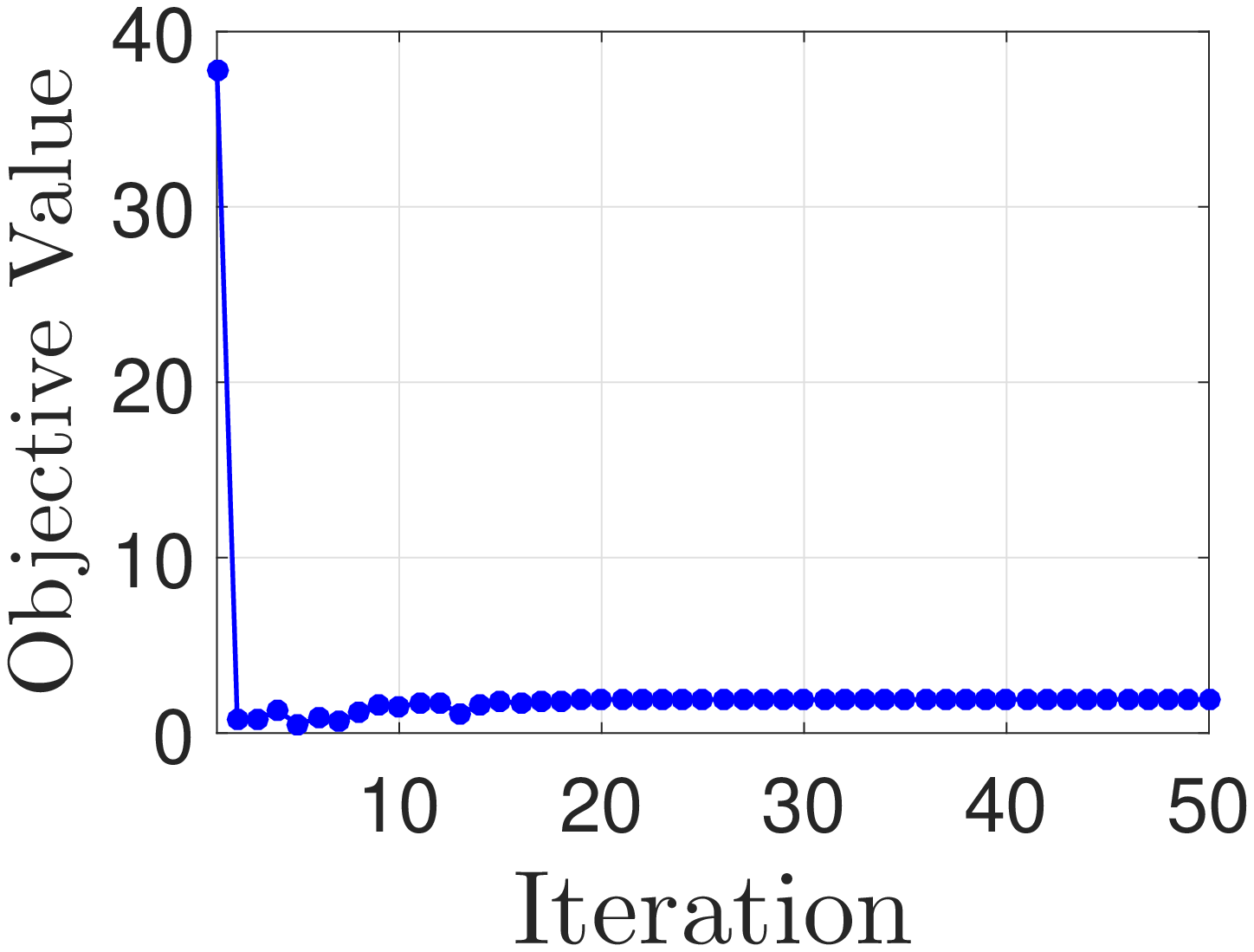}
		\caption{}		
	\end{subfigure}		
	\caption{(a) Average $IP$ value for UTKinect, Words, and DaynTex++. (b) Convergence curve of CKSC for Cricket. }
	\label{fig:conv}
\end{figure}
\subsection{Interpretability of the Dictionary}
We define the interpretability measure $IP_i$ for each $\na_i$ as 
$$IP_i={\max_{j}(\vec{\rho}^\top_j\na_i)}/(\vec{\MB{1}}^\top \nH \na_i)$$
where $\vec{\MB{1}}\in \mathbb{R}^{\ncl}$ is a vector of ones. 
$IP_i$ becomes $1$ if 
$\na_i$ uses data instances related only to one specific class.
Figure \ref{fig:conv}a presents the $IP$ value for the algorithms CKSC, EKDL, LC-NNKSC, KGDL, and LP-KSVD 
related to their implementations on the datasets Words, DynTex++, and UTKinect.
Based on the results, CKSC and LC-NNKSC achieved the highest $IP$ values as they use similar non-negative constraints in their training frameworks.
Among others, EKDL presents better interpretability results due to the incoherency term it uses between the dictionary atoms $\na_i$.
%
Putting the above results next to the classification accuracies (Table \ref{tab:kernel}), 
we conclude that the CKSC algorithm learns highly interpretable dictionary atoms $\na_i$ while providing an efficient discriminative representation.

%% file: conclusion.tex
\section{Conclusion}
In this work, we presented a novel kernel-based discriminant sparse coding and dictionary learning framework, 
which relies on the discriminative confidence of each sparse codes in an individual class of data.
Our CKSC algorithm incorporates the labeling information in training of the dictionary as well as in reconstruction of the test data in the kernel space, which increases the consistency between its trained and recall models.
It also benefits from a non-negative framework which facilitates the discriminant reconstruction of the data points using the contributions taken from the most similar class of data.
CKSC algorithm is presented and discussed comprehensively. 
Based on the empirical evaluations on time-series datasets, CKSC outperforms the relevant kernel-based sparse coding baselines due to higher classification performance based on the obtained representations. It also learns dictionaries with the more interpretable structures compared to other algorithms.  
%
%
%
%